%% file: main.tex
\documentclass[10pt,twocolumn,letterpaper]{article}

\usepackage{cvpr}              

\input{preamble}

\definecolor{cvprblue}{rgb}{0.21,0.49,0.74}
\usepackage[pagebackref,breaklinks,colorlinks,allcolors=cvprblue]{hyperref}

\def\paperID{*****} 
\def\confName{CVPR}
\def\confYear{2026}

\title{InpaintHuman: Reconstructing Occluded Humans with Multi-Scale UV Mapping and Identity-Preserving Diffusion Inpainting}

\author{
    Jinlong Fan$^{1}$ \quad
    Shanshan Zhao$^{2}$ \quad
    Liang Zheng$^{1}$ \quad
    Jing Zhang$^{3}$ \quad
    Yuxiang Yang$^{1,*}$ \quad
    Mingming Gong$^{4}$ \\[5pt]
    $^{1}$Hangzhou Dianzi University \quad
    $^{2}$Alibaba International Digital Commerce Group \\
    $^{3}$Wuhan University \quad
    $^{4}$University of Melbourne
}

\begin{document}

\twocolumn[{%
\renewcommand\twocolumn[1][]{#1}%
\maketitle
\includegraphics[width=1.0\linewidth]{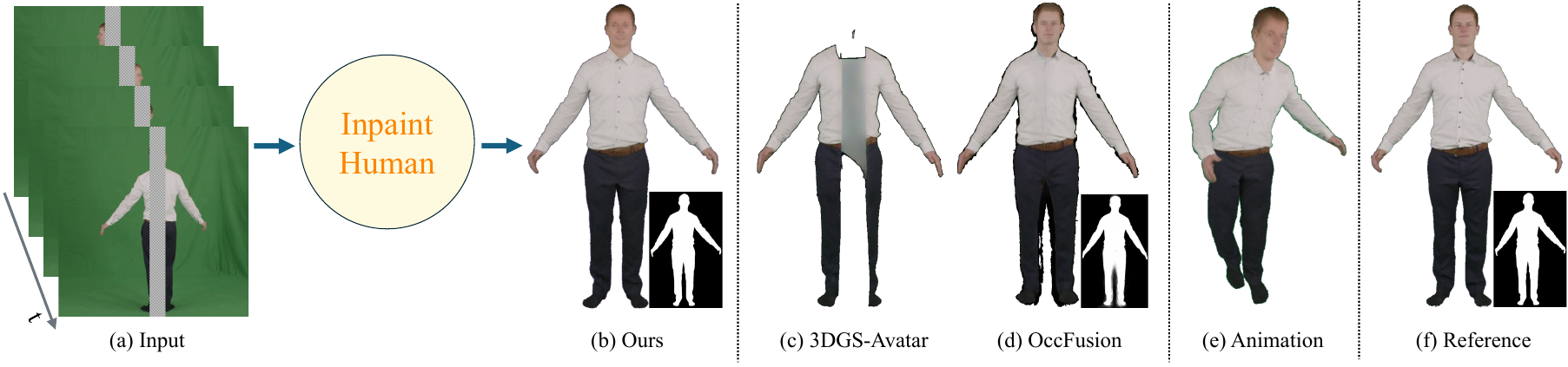}
\captionof{figure}{Given a video with significant occlusions (a), existing methods produce incomplete or inconsistent reconstructions (c,d). InpaintHuman leverages occlusion-robust multi-scale UV-parameterized representation and identity-preserving diffusion inpainting to reconstruct a complete, animatable avatar with consistent appearance across novel views and poses (b,e).
\vspace{1em}
}

\label{fig:teaser}
}]

\blfootnote{
    \hspace{-1.8em}$^*$Corresponding author.\\
    E-mail: {\tt\small \{jfan, zhlbsbx, yyx\}@hdu.edu.cn, \\ \{sshan.zhao00, jingzhang.cv\}@gmail.com, Mingming.gong@unimelb.edu.au}
}

\input{sec/0_abstract}    
\input{sec/1_intro}

\input{sec/2_Related_work}
\input{sec/3_Method}
\input{sec/4_Experiments}
\input{sec/5_Conclusion}

{
    \small
    \bibliographystyle{ieeenat_fullname}
    \bibliography{main}
}


\end{document}

%% file: preamble.tex
\newcommand\blfootnote[1]{%
  \begingroup
  \renewcommand\thefootnote{}\footnote{#1}%
  \addtocounter{footnote}{-1}%
  \endgroup
}

\usepackage{multirow}
\usepackage{booktabs}

%% file: sec/0_abstract.tex
\begin{abstract}
Reconstructing complete and animatable 3D human avatars from monocular videos remains challenging, particularly under severe occlusions. While 3D Gaussian Splatting has enabled photorealistic human rendering, existing methods struggle with incomplete observations, often producing corrupted geometry and temporal inconsistencies. We present InpaintHuman, a novel method for generating high-fidelity, complete, and animatable avatars from occluded monocular videos. Our approach introduces two key innovations: (i) a multi-scale UV-parameterized representation with hierarchical coarse-to-fine feature interpolation, enabling robust reconstruction of occluded regions while preserving geometric details; and (ii) an identity-preserving diffusion inpainting module that integrates textual inversion with semantic-conditioned guidance for subject-specific, temporally coherent completion. Unlike SDS-based methods, our approach employs direct pixel-level supervision to ensure identity fidelity. Experiments on synthetic benchmarks (PeopleSnapshot, ZJU-MoCap) and real-world scenarios (OcMotion) demonstrate competitive performance with consistent improvements in reconstruction quality across diverse poses and viewpoints.
\end{abstract}

%% file: sec/1_intro.tex
\section{Introduction}
\label{sec:intro}
Reconstructing animatable 3D human avatars from monocular videos is essential for applications in virtual reality, augmented reality, telepresence, and digital content creation. Recent advances in neural rendering, particularly Neural Radiance Fields (NeRF)~\cite{mildenhall2021nerf}, and 3D Gaussian Splatting (3DGS)~\cite{kerbl20233d}, have achieved impressive results in capturing photorealistic human appearances. However, these methods typically assume full visibility of the target human throughout the input sequence, a condition rarely satisfied in practice. In real-world environments, occlusions caused by other individuals, environmental elements, or self-occlusion frequently lead to incomplete geometry, degraded texture fidelity, and temporal inconsistencies.

Two fundamental limitations hinder robust human reconstruction under occlusions. First, existing methods such as HumanNeRF~\cite{weng2022humannerf} and GaussianAvatar~\cite{hu2024gaussianavatar} optimize scene-specific representations that lack the capacity to hallucinate unseen regions without ground-truth supervision, resulting in holes and visual artifacts in occluded areas. Second, recent occlusion-aware approaches like OccNeRF~\cite{xiang2023rendering} rely primarily on interpolating observed visual cues to infer unseen parts. While effective for minor occlusions, these techniques struggle to generate plausible appearances for extensively or completely unobserved body regions.

The emergence of generative diffusion models~\cite{ho2020denoising,rombach2022high} presents a promising avenue for synthesizing missing content. Prior efforts integrating diffusion priors with 3D representations through Score Distillation Sampling (SDS)~\cite{poole2022dreamfusion} have demonstrated compelling results in static scene completion. However, applying such techniques to dynamic human reconstruction introduces critical challenges: (i) \emph{identity drift}, where stochastic variations in diffusion sampling cause inconsistent appearance across frames, and (ii) \emph{supervision ambiguity}, arising from the indirect nature of gradient-based diffusion guidance, which hampers precise geometry optimization.

To address these challenges, we propose \textbf{InpaintHuman}, a diffusion-enhanced reconstruction method for occluded human avatar generation. Our approach introduces two core innovations. First, we develop a \emph{multi-scale UV-parameterized representation} that operates in canonical pose space, providing inherent robustness against occlusions through hierarchical coarse-to-fine feature interpolation while preserving fine-grained geometric and textural details. Second, we design an \emph{identity-preserving diffusion inpainting module} that ensures subject-specific and temporally coherent completion of unseen body parts by leveraging textual inversion~\cite{gal2022image} to capture concept-level identity characteristics and employing semantics-guided personalized diffusion inpainting.

Notably, unlike SDS-based methods that rely on latent-space supervision with inherent stochasticity, our approach leverages \emph{direct pixel-level supervision} in image space. The reconstruction pipeline proceeds as follows: we first initialize 3D Gaussians in canonical space based on visible observations, then train a subject-specific inpainting model to synthesize complete and identity-consistent textures, and subsequently refine the Gaussian field using the inpainted results as supervision.

We conduct extensive evaluations on synthetic occlusion benchmarks, including PeopleSnapshot~\cite{alldieck2018video} and ZJU-MoCap~\cite{peng2021neural}, as well as real-world scenarios from OcMotion~\cite{huang2022occluded}. Experimental results demonstrate that InpaintHuman achieves consistent improvements in both visible-region fidelity and plausibility of reconstructed occluded areas. Our main contributions are:

\begin{itemize}[leftmargin=*,nosep]
    \item We present InpaintHuman, a novel method for reconstructing complete, animatable 3D human avatars from occluded monocular videos.
    
    \item We propose a multi-scale UV-parameterized representation that enables robust occlusion handling through hierarchical coarse-to-fine feature interpolation while maintaining fine geometric details.
    
    \item We introduce an identity-preserving diffusion inpainting strategy combining textual inversion with semantic-conditioned guidance for subject-specific, temporally coherent completion of occluded body parts.
\end{itemize}

%% file: sec/2_Related_work.tex
\section{Related Work}
\label{sec:related}

\subsection{3D Human Avatar Reconstruction}

Neural rendering has revolutionized human avatar reconstruction from monocular video. NeRF-based methods such as Neural Body~\cite{peng2021neural} and HumanNeRF~\cite{weng2022humannerf} achieve high-fidelity rendering by encoding human appearance in neural radiance fields conditioned on body pose. However, these approaches suffer from slow rendering and sensitivity to pose estimation errors~\cite{guo2023vid2avatar,jiang2022neuman,jiang2022selfrecon,yu2023monohuman,sun2023neural}. More recently, 3D Gaussian Splatting~\cite{kerbl20233d} has emerged as an efficient alternative, enabling real-time rendering with explicit geometry. Methods like GauHuman~\cite{hu2024gauhuman}, 3DGS-Avatar~\cite{qian20243dgs}, and GaussianAvatar~\cite{hu2024gaussianavatar} extend this representation to dynamic humans by anchoring Gaussians on parametric body models. While these approaches achieve impressive results under full visibility, they fundamentally lack mechanisms to handle missing observations, leading to degraded performance under occlusion~\cite{kocabas2024hugs,moreau2024human,liu2024animatable,pang2024ash,li2024gaussianbody}. Our work builds upon Gaussian-based representations but specifically addresses the occlusion challenge through multi-scale UV parameterization and diffusion-guided completion.

\subsection{Occlusion-Aware Human Reconstruction}

Handling occlusions in human reconstruction has received increasing attention. OccNeRF~\cite{xiang2023rendering} introduces surface-based rendering with geometry and visibility priors to improve robustness, but remains limited by its reliance on observed data for inferring unseen regions. OccGaussian~\cite{ye2025occgaussian} extends Gaussian splatting with occlusion-aware training strategies. Wild2Avatar~\cite{xiang2023wild2avatar} tackles in-the-wild scenarios but struggles with severe occlusions. More recent approaches leverage generative priors: OccFusion~\cite{sun2024occfusion} and Guess The Unseen (GTU)~\cite{lee2024guess} integrate diffusion models through SDS-based optimization, while WonderHuman~\cite{wang2025wonderhuman} employs multi-view diffusion priors. However, SDS-based methods commonly encounter identity drift due to stochastic sampling and supervision ambiguity from indirect gradient flow. Our approach mitigates these issues by training a personalized inpainting model that provides direct pixel-level supervision with identity-consistent completion.

\subsection{Diffusion Models for Image Inpainting}

Diffusion models~\cite{ho2020denoising,songscore} have demonstrated remarkable capabilities in image generation and editing. Stable Diffusion~\cite{rombach2022high} enables efficient high-resolution synthesis through latent-space diffusion. For inpainting tasks, models such as Stable Diffusion Inpainting~\cite{rombach2022high} and SDXL-Inpainting~\cite{podell2023sdxl} achieve impressive results by conditioning on masked images. To enable subject-specific generation, textual inversion~\cite{gal2022image} and DreamBooth~\cite{ruiz2023dreambooth} learn personalized embeddings from few-shot examples. ControlNet~\cite{zhang2023adding} provides spatial conditioning through auxiliary inputs such as pose or depth maps~\cite{mou2024t2i}. 
We leverage these advances by combining textual inversion for identity preservation with ControlNet for pose consistency, trained in a self-supervised manner on visible regions to achieve subject-specific inpainting.

%% file: sec/3_Method.tex
\section{Method}
\label{sec:method}

\subsection{Overview}
\label{sec:overview}

Given a monocular video of an occluded human, our goal is to reconstruct a complete and animatable 3D avatar with high-fidelity appearance and temporal consistency. For each frame $I_i \in \{I_1, \ldots, I_N\}$, we utilize SMPL~\cite{loper2023smpl} parameters $(\beta, \theta_i)$ and a visibility mask $\mathcal{M}_{\text{vis}}^i$ indicating observed body regions. The central challenge lies in synthesizing plausible geometry and texture for unobserved regions while preserving subject-specific identity across varying poses.

\begin{figure*}
    \centering
    \includegraphics[width=0.95\linewidth]{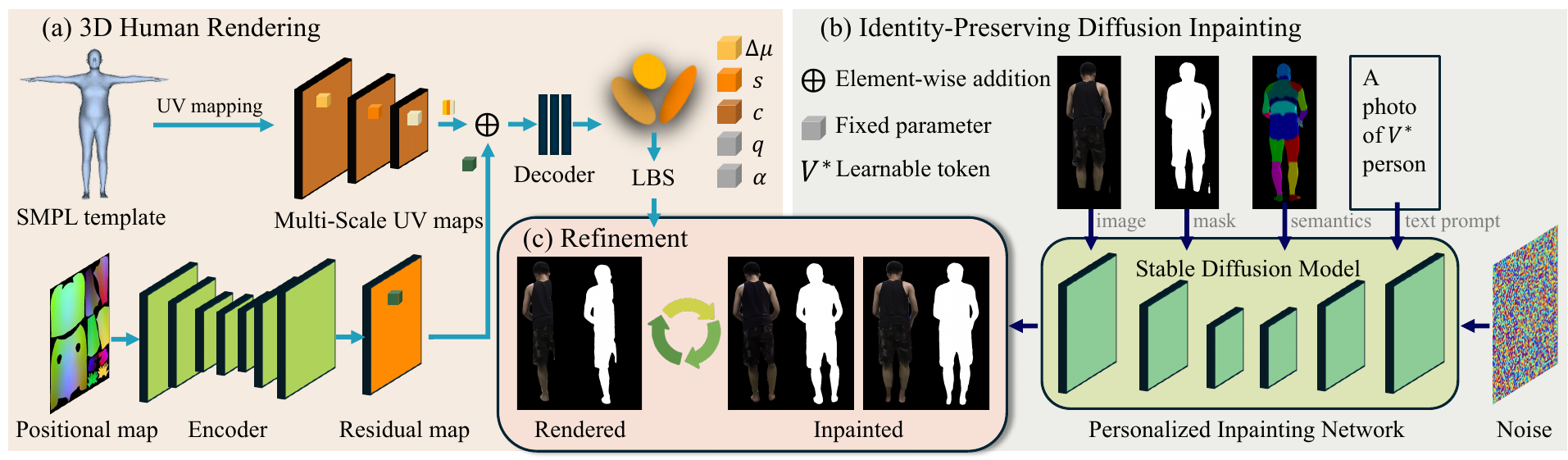}
    \caption{\textbf{Overview of the InpaintHuman.} 
    \textbf{(a) 3D Human Rendering:} We represent the human avatar using 3D Gaussians anchored on the SMPL mesh, with attributes predicted from multi-scale UV feature maps that enable robust interpolation across occluded regions. These Gaussians are transformed to observation space via forward LBS, augmented with pose-dependent residual features for non-rigid dynamics. 
    \textbf{(b) Identity-Preserving Diffusion Inpainting:} A personalized Stable Diffusion inpainting model takes occluded images and visibility masks as input. Subject-level identity is captured via textual inversion with a learnable token, while pose consistency is ensured through ControlNet-based semantic guidance.
    \textbf{(c) Refinement:} Inpainted images supervise the optimization of canonical UV maps, propagating plausible content to occluded regions and yielding a complete, animatable avatar.
    }
    \label{fig:pipeline}
\end{figure*}

Our approach addresses this challenge through two synergistic components. First, we represent the avatar using a \emph{multi-scale UV-parameterized canonical representation} (Sec.~\ref{sec:representation}), which encodes appearance in a pose-independent space and enables robust feature interpolation for occluded regions. Second, we introduce an \emph{identity-preserving diffusion inpainting module} (Sec.~\ref{sec:inpainting}) that leverages personalized generative priors to synthesize complete, subject-specific textures. These inpainted results serve as pixel-level supervision to \emph{refine the canonical representation} (Sec.~\ref{sec:training}), yielding a coherent and animatable avatar. An overview is illustrated in Fig.~\ref{fig:pipeline}.

\subsection{3D Human Rendering}
\label{sec:representation}

\subsubsection{Canonical Space Representation}
\paragraph{Template-Based Canonicalization.}
To establish a pose-independent representation, we leverage the SMPL body model~\cite{loper2023smpl} parameterized by shape $\beta$ and pose $\theta$. We define the canonical space using a rest pose $\theta_0$ (A-pose) and initialize $N$ points $\{x_i\}_{i=1}^{N}$ by sampling on the template mesh surface. For each point, we compute blend skinning weights $\mathbf{w}_i \in \mathbb{R}^{K}$ via barycentric interpolation, where $K$ denotes the number of joints.

A key advantage of using the SMPL template is its predefined UV parameterization, which maps each 3D point $x_i \in \mathbb{R}^3$ to 2D coordinates $(u_i, v_i) \in [0, 1]^2$. This unwrapping enables us to represent the human surface as a 2D manifold, facilitating efficient feature manipulation through convolutional operations.

Each sampled point $x_i$ is associated with a 3D Gaussian primitive~\cite{kerbl20233d} characterized by: center position $\mu_i \in \mathbb{R}^3$, color $c_i \in \mathbb{R}^3$, opacity $\alpha_i \in \mathbb{R}$, rotation quaternion $q_i \in \mathbb{R}^4$, and scale $s_i \in \mathbb{R}^3$. Following prior work~\cite{hu2024gaussianavatar}, we adopt simplifications to enhance training stability in the monocular setting: (i) fixing opacity to $\alpha = 1$, (ii) using isotropic Gaussians with scalar scale $s \in \mathbb{R}$, and (iii) initializing rotation to the identity quaternion $q = (1, 0, 0, 0)$.

\paragraph{Multi-Scale UV Mapping.}
Representing the 3D human as 2D UV feature maps offers distinct advantages for handling occlusions. In UV space, neighboring pixels correspond to adjacent points on the body surface, preserving semantic locality. In contrast, 3D Euclidean proximity can be misleading. For instance, points on the chest may be spatially closer to the upper arm than to adjacent chest regions, leading to erroneous cross-part interpolation.

\begin{figure}
    \centering
    \includegraphics[width=1.0\linewidth]{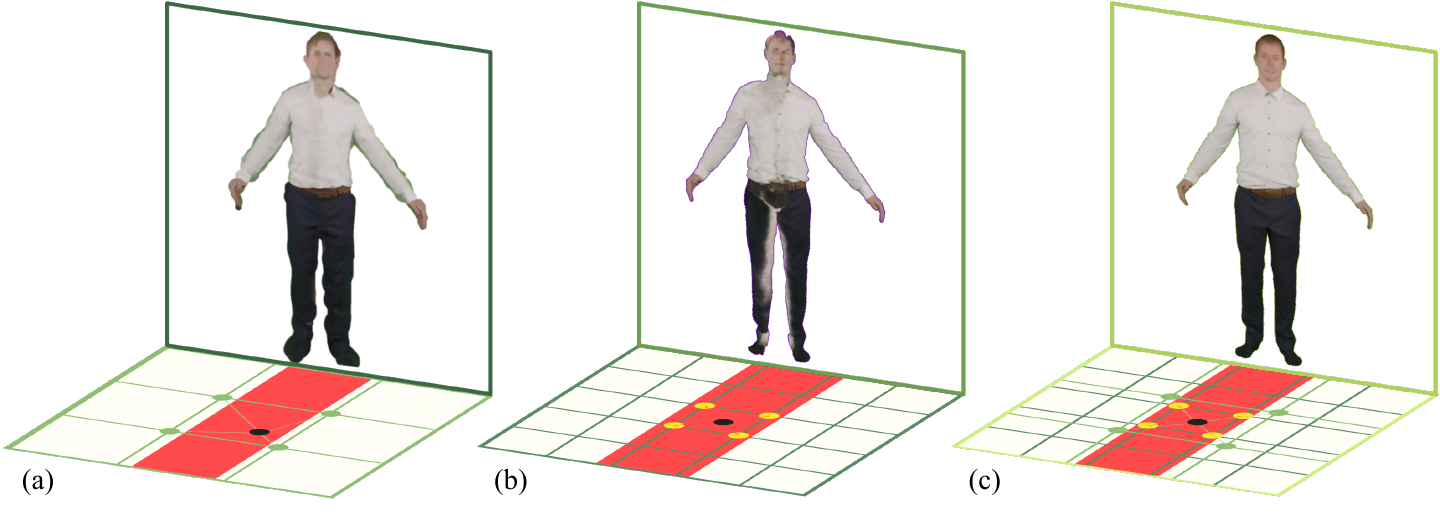}
    \caption{\textbf{Multi-scale UV feature maps for occlusion robustness.} Coarser resolutions (e.g., $64 \times 64$) compress spatial distances between visible and occluded regions, facilitating feature interpolation but lacking fine details. Higher resolutions (e.g., $256 \times 256$) preserve geometric details but are more susceptible to incomplete observations. Our hierarchical design combines both advantages: robust occlusion handling with high-fidelity detail preservation.}
    \label{fig:multi-scale-feature}
\end{figure}

We observe a fundamental trade-off in UV map resolution: coarser maps compress spatial distance between visible and occluded regions, facilitating feature propagation but sacrificing fine details; finer maps capture high-frequency geometry but are more susceptible to incomplete observations. To leverage both strengths, we construct a hierarchy of UV feature maps $\{\mathcal{F}_l\}_{l=1}^{L}$ at $L$ different resolutions ($64 \times 64$, $128 \times 128$, and $256 \times 256$ in our implementation). As illustrated in Fig.~\ref{fig:multi-scale-feature}, coarser maps provide robustness to occlusions through effective spatial interpolation, while finer maps preserve geometric details for high-fidelity rendering.

\paragraph{Gaussian Parameter Decoder.}  
Given a 3D point $x_i$ with UV coordinates $(u_i, v_i)$, we sample features from each scale using bilinear interpolation: $f_l = \mathcal{F}_l(u_i, v_i)$. The multi-scale features are aggregated via summation to obtain the canonical feature $f_c = \sum_{l=1}^{L} f_l$. This feature, concatenated with positional encoding $\gamma(\mu_i)$, is fed into a lightweight MLP decoder $\mathcal{D}$ to predict Gaussian attributes:
\begin{equation}
    (\Delta\mu_i, s_i, c_i) = \mathcal{D}(f_c, \gamma(\mu_i)),
\end{equation}
where $\Delta\mu_i$ denotes the position offset from the template surface, $s_i$ is the isotropic scale, and $c_i \in \mathbb{R}^3$ is the RGB color.

\subsubsection{Dynamics Modeling}
To render the avatar in a target pose, the canonical representation must be transformed to observation space. We decompose human dynamics into rigid articulation via Linear Blend Skinning (LBS) and non-rigid deformations via pose-dependent residual features.

\paragraph{Rigid Transformation.}
Given target pose $\theta_t$, each Gaussian center is transformed from canonical to posed space using forward LBS. Let $\mathbf{T}_k(\theta_t) \in SE(3)$ denote the transformation matrix for joint $k$. The posed position $\mu_i^t$ is computed as:
\begin{equation}
    \mu_i^t = \left( \sum_{k=1}^{K} w_{i,k} \mathbf{T}_k(\theta_t) \right) \bar{\mu}_i,
\end{equation}
where $\bar{\mu}_i = \mu_i + \Delta\mu_i$ is the canonical position with predicted offset.

\paragraph{Pose-Dependent Residual Features.}
While LBS captures skeletal motion, it cannot model pose-dependent appearance variations such as clothing wrinkles. We render the SMPL mesh at pose $\theta_t$ into a position map $\mathcal{P}_t \in \mathbb{R}^{H \times W \times 3}$, which is processed by a convolutional encoder $\mathcal{E}$ to produce a residual feature map $\mathcal{R}_t = \mathcal{E}(\mathcal{P}_t)$. For each point, we sample its residual feature $f_t = \mathcal{R}_t(u_i, v_i)$ and combine it with the canonical feature: $f = f_c + f_t$. The combined feature is decoded via Eq.~(1) to predict pose-specific Gaussian attributes. Finally, the posed Gaussians are rendered using tile-based rasterization~\cite{kerbl20233d}.

\subsection{Identity-Preserving Diffusion Inpainting}
\label{sec:inpainting}

While our multi-scale UV representation enables robust feature interpolation for partially occluded regions, it cannot hallucinate plausible content for body parts that are never observed throughout the entire video sequence. To address this limitation, we leverage pre-trained diffusion models to synthesize complete appearances. However, directly applying off-the-shelf inpainting leads to \emph{identity drift} that generated content may be realistic but inconsistent with the subject's actual appearance.

To tackle this challenge, our diffusion inpainting module operates at two complementary levels. At the \emph{subject level}, we employ textual inversion to learn a global token that captures the individual's distinctive characteristics, such as clothing style and overall appearance. At the \emph{pose level}, we incorporate semantic guidance through ControlNet to ensure that generated content respects the underlying body structure and remains spatially coherent across different poses. Together, these two components enable our model to produce completions that are both identity-consistent and anatomically correct.

\paragraph{Subject-Level Tokenization via Textual Inversion.}

Standard text-to-image diffusion models, trained on generic image-caption pairs, lack such subject-specific knowledge and therefore cannot reliably generate content that matches a particular person. To bridge this gap, we employ textual inversion~\cite{gal2022image} to learn a dedicated token $V^*$ that encapsulates the identity characteristics of the target individual. Specifically, given a collection of visible (non-occluded) frames $\{I_i^{\text{vis}}\}$ extracted from the input video, we optimize a learnable embedding $v^* \in \mathbb{R}^{d}$ that maps to the token $V^*$ in the text encoder's vocabulary. The optimization encourages the diffusion model to faithfully reconstruct the visible content when conditioned on prompts containing this learned token:
\begin{equation}
    \mathcal{L}_{\text{TI}} = \mathbb{E}_{z, \epsilon, t} \left[ \| \epsilon - \epsilon_\phi(z_t, t, \tau_\psi(V^*)) \|_2^2 \right],
\end{equation}
where $z_t$ denotes the noised latent representation at diffusion timestep $t$, $\epsilon_\phi$ is the denoising network, and $\tau_\psi$ is the text encoder. We jointly fine-tune $\tau_\psi$ along with the learnable embedding $v^*$. Once learned, the token $V^*$ serves as a compact yet powerful representation of the subject's identity. When incorporated into generation prompts (e.g., ``a photo of $V^*$''), it guides the diffusion model to produce outputs that remain visually coherent with the individual's distinctive traits. 

\paragraph{Semantic-Guided Personalized Inpainting.}

To enforce pose consistency, we incorporate ControlNet~\cite{zhang2023adding} with semantic conditioning derived from the SMPL body model. For each frame, we render a semantic map $\mathcal{S}_t$ from the fitted SMPL mesh, which encodes body part labels and spatial layout information. ControlNet then injects this semantic guidance into the diffusion process, ensuring that generated content adheres to the correct body configuration. This conditioning is particularly important for maintaining temporal coherence. Without it, inpainted regions might exhibit inconsistent structures when the subject moves between poses.

A critical component of our approach is the self-supervised training strategy, which enables the model to learn appearance priors directly from the input video without requiring external supervision. For each training iteration, we sample a frame $I_t$ along with its visibility mask $\mathcal{M}_{\text{vis}}$, and then apply an additional random mask $\mathcal{M}_{\text{rand}}$ to the visible regions: $\mathcal{M}_{\text{train}} = \mathcal{M}_{\text{vis}} \odot \mathcal{M}_{\text{rand}}$.

The model is trained to inpaint these randomly masked visible pixels, with ground truth readily available from the original frame. This self-supervised objective serves a dual purpose: it teaches the model to extract and propagate appearance features from observed regions, while simultaneously adapting the generic diffusion prior to the specific visual characteristics of the target subject.
To maintain training efficiency while enabling effective adaptation, we employ Low-Rank Adaptation (LoRA)~\cite{hulora} to fine-tune the pre-trained Stable Diffusion inpainting model~\cite{rombach2022high}. The overall training objective combines diffusion denoising with identity and pose conditioning:
\begin{equation}
    \mathcal{L}_{\text{inpaint}} = \mathbb{E}_{z, \epsilon, t, \mathcal{S}} \left[ \| \epsilon - \epsilon_\phi(z_t, t, \tau_\psi(V^*), \mathcal{C}(\mathcal{S})) \|_2^2 \right],
\end{equation}
where $\mathcal{C}(\mathcal{S})$ denotes the ControlNet conditioning derived from the semantic map. At inference, the model generates complete human image $\tilde{I}_t$, and SAM~\cite{ke2023segment} produces complete mask $\mathcal{M}_{\text{full}}$ for supervising canonical refinement.

\subsection{Training Strategy and Objective}
\label{sec:training}

Our training consists of three progressive stages (Fig.~\ref{fig:pipeline}(c)).

\paragraph{Stage 1: Canonical Initialization.}
We optimize multi-scale UV feature maps and decoder using visible regions:
\begin{equation}
    \mathcal{L}_{\text{init}} = \sum_{p \in \mathcal{M}_{\text{vis}}} \| I(p) - \hat{I}(p) \|_1.
\end{equation}

\paragraph{Stage 2: Diffusion Model Personalization.}
We fine-tune the diffusion inpainting model via $\mathcal{L}_{\text{inpaint}}$ (Sec.~\ref{sec:inpainting}) to learn identity-consistent completions.

\paragraph{Stage 3: Canonical Refinement.}
We refine canonical UV maps using inpainted images as pseudo ground truth:
\begin{equation}
    \mathcal{L}_{\text{refine}} = \sum_{p \in \mathcal{M}_{\text{full}}} \left( \| \tilde{I}(p) - \hat{I}(p) \|_1 + \lambda_{\text{ssim}} \mathcal{L}_{\text{ssim}} + \lambda_{\text{lpips}} \mathcal{L}_{\text{lpips}} \right).
\end{equation}
The total objective is $\mathcal{L}_{\text{total}} = \mathcal{L}_{\text{init}} + \lambda_{\text{refine}} \mathcal{L}_{\text{refine}}$.

\begin{table*}[t]
    \centering
    
    \begin{tabular}{l|ccc|ccc}
        \toprule
        \multirow{2}{*}{Method} & \multicolumn{3}{c|}{ZJU-MoCap~\cite{peng2021neural}} & \multicolumn{3}{c}{OcMotion~\cite{huang2022occluded}} \\
        \cline{2-7}
         & PSNR$\uparrow$ & SSIM$\uparrow$ & LPIPS*$\downarrow$ & PSNR$\uparrow$ & SSIM$\uparrow$ & LPIPS*$\downarrow$ \\
        \midrule
        HumanNeRF~\cite{weng2022humannerf}       & 20.67 & 0.9509 & --    & 9.79  & 0.7203 & 189.1 \\
        3DGS-Avatar~\cite{qian20243dgs}   & 17.29 & 0.9410 & 63.25 & --    & --     & --    \\
        GauHuman~\cite{hu2024gauhuman}         & 21.55 & 0.9430 & 55.88 & 15.09 & 0.8525 & 107.1 \\
        GaussianAvatar~\cite{hu2024gaussianavatar} & 18.01 & 0.9512 & 60.33 & --    & --     & --    \\
        \midrule
        OccNeRF~\cite{xiang2023rendering}           & 22.40 & 0.9562 & 43.01 & 15.71 & 0.8230 & 82.90 \\
        OccGaussian~\cite{ye2025occgaussian}   & 23.29 & 0.9482 & 41.93 & --    & --     & --    \\
        Wild2Avatar~\cite{xiang2023wild2avatar}   & --    & --     & --    & 14.09 & 0.8484 & 93.21 \\
        GTU~\cite{lee2024guess}                   & 22.89 & 0.9503 & 40.78 & 15.83 & 0.8437 & 83.46 \\
        OccFusion~\cite{sun2024occfusion}       & \underline{23.96} & \underline{0.9548} & \underline{32.34} & \underline{18.28} & \underline{0.8805} & \underline{82.42} \\
        \midrule
        InpaintHuman (Ours)              & \textbf{24.65} & \textbf{0.9614} & \textbf{31.63} & \textbf{19.02} & \textbf{0.8946} & \textbf{81.98} \\
        \bottomrule
    \end{tabular}

    \caption{\textbf{Quantitative comparison on ZJU-MoCap~\cite{peng2021neural} and OcMotion~\cite{huang2022occluded} datasets.} Methods in the upper section are standard human rendering approaches, while those in the lower section are designed for occluded scenarios. ``--'' indicates results not available. The \textbf{best} and \underline{second-best} results are highlighted. On both datasets, InpaintHuman achieves competitive or superior performance, demonstrating the effectiveness of our identity-preserving approach.}
    \label{tab:quantitative_results}
    
\end{table*}

%% file: sec/4_Experiments.tex
\section{Experiments}
\label{sec:experiments}

\begin{figure*}[htbp!]
    \centering
    \includegraphics[width=0.85\linewidth]{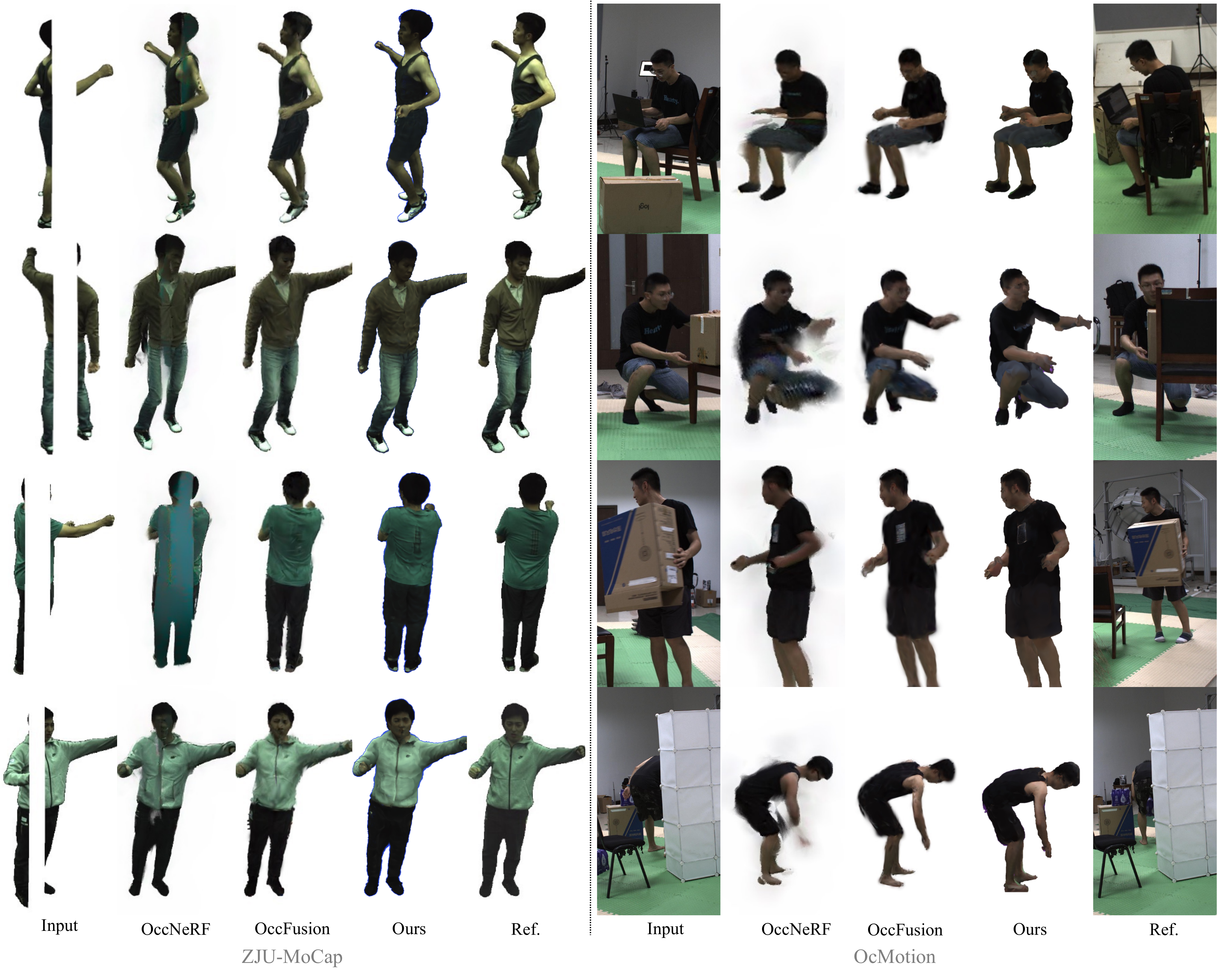}
    \caption{\textbf{Qualitative comparison on novel view synthesis.} We present results on ZJU-MoCap~\cite{peng2021neural} with synthetic occlusions (left) and OcMotion~\cite{huang2022occluded} with real-world occlusions (right). OccNeRF~\cite{xiang2023rendering} struggles to hallucinate unseen regions, often producing noticeable discoloration. OccFusion~\cite{sun2024occfusion} generates sharper textures in some areas but exhibits blurriness and visual uncertainty in heavily occluded regions. Our method produces more complete renderings with better preservation of subject-specific appearance.}
    \label{fig:visual_results}
\end{figure*}

We evaluate InpaintHuman on both synthetic and real-world occlusion scenarios, comparing against state-of-the-art methods and validating our design choices through ablation studies.

\subsection{Experimental Setup}
\label{sec:setup}

We conduct experiments on three datasets. \textbf{PeopleSnapshot}~\cite{alldieck2018video} contains monocular videos of individuals rotating before a stationary camera; we synthesize occlusions for controlled evaluation. \textbf{ZJU-MoCap}~\cite{peng2021neural} consists of 6 dynamic subjects captured by a synchronized multi-camera system. Following OccNeRF~\cite{xiang2023rendering}, we mask the central 50\% of human pixels for the first 80\% of frames, sample 100 frames at intervals of 5 from the first camera for training, and use remaining 22 views for evaluation. \textbf{OcMotion}~\cite{huang2022occluded} comprises 48 videos with naturally occurring occlusions from human-object interactions. Following OccFusion~\cite{sun2024occfusion}, we evaluate on 6 diverse sequences with 50 subsampled frames each.

\paragraph{Baselines.}
We compare against two categories of methods: (1) \emph{standard human rendering methods} not specifically designed for occlusion, including HumanNeRF~\cite{weng2022humannerf}, 3DGS-Avatar~\cite{qian20243dgs}, GauHuman~\cite{hu2024gauhuman}, and GaussianAvatar~\cite{hu2024gaussianavatar}; and (2) \emph{occlusion-aware approaches}, including OccNeRF~\cite{xiang2023rendering}, OccGaussian~\cite{ye2025occgaussian}, Wild2Avatar~\cite{xiang2023wild2avatar}, OccFusion~\cite{sun2024occfusion}, and Guess The Unseen (GTU)~\cite{lee2024guess}.  For fair comparison, all methods use identical segmentation masks and pose priors.

\paragraph{Metrics.}
We report PSNR, SSIM~\cite{wang2004image}, and LPIPS~\cite{zhang2018unreasonable} (reported as LPIPS* = 1000 $\times$ LPIPS for clarity). Since OcMotion lacks ground truth for occluded regions, metrics are computed over visible pixels only.

\subsection{Implementation Details}
\label{sec:implementation}

The multi-scale UV feature maps are set to resolutions of $64 \times 64$, $128 \times 128$, and $256 \times 256$, with feature dimensions of 32 per scale. The Gaussian parameter decoder is a 3-layer MLP with hidden dimension 128. For diffusion inpainting, we use Stable Diffusion v2 Inpainting as the backbone, with LoRA rank set to 8. We use the AdamW optimizer with learning rate $1 \times 10^{-4}$ for the canonical representation and $1 \times 10^{-5}$ for LoRA parameters. The loss weights are set to $\lambda_{\text{ssim}} = 0.2$, $\lambda_{\text{lpips}} = 0.1$, and $\lambda_{\text{refine}} = 1.0$. Training takes approximately 40 minutes on a single NVIDIA RTX 4090 GPU.

\begin{figure*}[htbp!]
    \centering
    \includegraphics[width=1.0\linewidth]{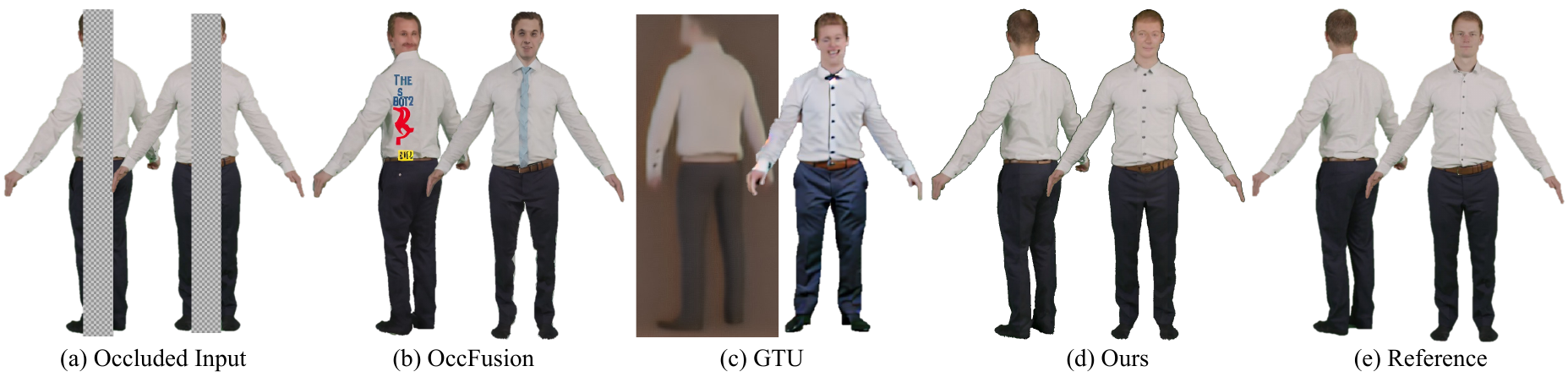}
    \caption{\textbf{Qualitative comparison of inpainting results.} Given occluded input images (a), we compare completions from OccFusion~\cite{sun2024occfusion} (b), GTU~\cite{lee2024guess} (c), and our method (d), with ground truth reference (e). Our identity-preserving diffusion module generates textures that maintain appearance consistency with visible regions and spatial plausibility respecting body structure. In contrast, SDS-based methods (b, c) exhibit identity drift with inconsistent colors and patterns.}
    \label{fig:inpaint}
\end{figure*}

\subsection{Evaluation Results}
\label{sec:results}

\subsubsection{Quantitative Results}

Table~\ref{tab:quantitative_results} summarizes quantitative comparisons on ZJU-MoCap~\cite{peng2021neural} and OcMotion~\cite{huang2022occluded} datasets. Several observations can be made from these results. First, methods specifically designed for occluded human rendering generally outperform standard approaches, as the latter lack explicit mechanisms to handle missing observations and thus suffer from degraded performance in occluded regions. Second, among occlusion-aware methods, InpaintHuman achieves competitive or superior performance across both datasets. On ZJU-MoCap, our method attains the highest metrics, outperforming both interpolation-based approaches (OccNeRF, OccGaussian) and SDS-based methods (GTU, OccFusion). On OcMotion with real-world occlusions, InpaintHuman also demonstrates favorable results, suggesting that our identity-preserving inpainting strategy generalizes well to challenging in-the-wild scenarios.

\subsubsection{Inpainting Quality}

Figure~\ref{fig:inpaint} illustrates the inpainting results produced by our identity-preserving diffusion module. The personalized model, conditioned on the learned subject token $V^*$ and pose guidance from ControlNet, generates textures that exhibit three desirable properties: (1) \emph{appearance consistency}, the inpainted regions maintain coherent color and texture patterns with the visible parts; (2) \emph{spatial plausibility}, the generated content respects body structure and anatomical constraints; and (3) \emph{temporal stability}, the completions remain consistent across different poses within the same sequence. These high-quality inpainted images subsequently serve as effective supervision for refining the canonical UV feature maps, enabling the reconstruction of complete human avatars from heavily occluded inputs.

\subsubsection{Rendering Quality}

Figure~\ref{fig:visual_results} presents qualitative comparisons on novel view synthesis. On ZJU-MoCap with synthetic occlusions (left), OccNeRF struggles to hallucinate content for unseen regions, often producing visible artifacts such as discoloration and floaters. In contrast, InpaintHuman generates more complete and identity-consistent renderings, benefiting from the direct pixel-level supervision provided by our personalized inpainting module.

On OcMotion with real-world occlusions (right), the challenges are more pronounced due to complex object interactions and diverse occlusion patterns. OccFusion, leveraging SDS-based optimization and in-context inpainting, generates sharper textures in some areas but exhibits blurriness and visual uncertainty in heavily occluded regions. While all methods show some degradation compared to synthetic scenarios, InpaintHuman maintains relatively stable performance, producing renderings with fewer artifacts and better preservation of subject identity. These results suggest that our approach offers improved robustness to realistic occlusion conditions encountered in practical applications.

\subsection{Ablation Studies}
\label{sec:ablation}

We conduct ablation studies on the PeopleSnapshot~\cite{alldieck2018video} sequence with synthetic occlusions to validate the contribution of each proposed component. Specifically, we evaluate: (1) multi-scale UV feature maps (MS Maps), (2) textual inversion for subject-level tokenization (TI), and (3) Semantic-based ControlNet guidance (SG).

\paragraph{Quantitative Analysis.}
Table~\ref{tab:ablation} reports the ablation results. The baseline without any proposed component achieves a PSNR of 20.05 dB, as it lacks effective mechanisms for handling occluded regions. Adding multi-scale UV maps improves PSNR to 22.35 dB (+2.30 dB), demonstrating the benefit of hierarchical feature interpolation for propagating information to partially occluded areas. Incorporating textual inversion further boosts performance to 24.27 dB (+1.92 dB), indicating that subject-level identity guidance is crucial for generating appearance-consistent completions. Finally, adding semantic guidance yields modest but consistent improvements across all metrics (PSNR: 24.31 dB, SSIM: 0.9701), suggesting that explicit pose conditioning helps maintain spatial coherence.

\begin{table}[t]
    \centering
    
    \resizebox{1.0\linewidth}{!}{
    \begin{tabular}{ccc|ccc}
        \toprule
        MS Maps & TI & SG & PSNR$\uparrow$ & SSIM$\uparrow$ & LPIPS*$\downarrow$ \\
        \midrule
                 &              &              & 20.05 & 0.9501 & 61.47 \\
        \checkmark &              &              & 22.35 & 0.9603 & 55.92 \\
        \checkmark & \checkmark &              & 24.27 & 0.9649 & 38.53 \\
        \checkmark & \checkmark & \checkmark & \textbf{24.31} & \textbf{0.9701} & \textbf{37.42} \\
        \bottomrule
    \end{tabular}
    }
    \caption{\textbf{Ablation study on PeopleSnapshot with synthetic occlusions.} We progressively add each proposed component to evaluate its contribution. MS: multi-scale UV feature maps; TI: textual inversion for subject-level tokenization; SG: semantic guidance via ControlNet. Each component provides consistent improvements, with the full model achieving the best performance across all metrics.}
    \label{tab:ablation}
\end{table}

\begin{figure}[t]
    \centering
    \includegraphics[width=0.9\linewidth]{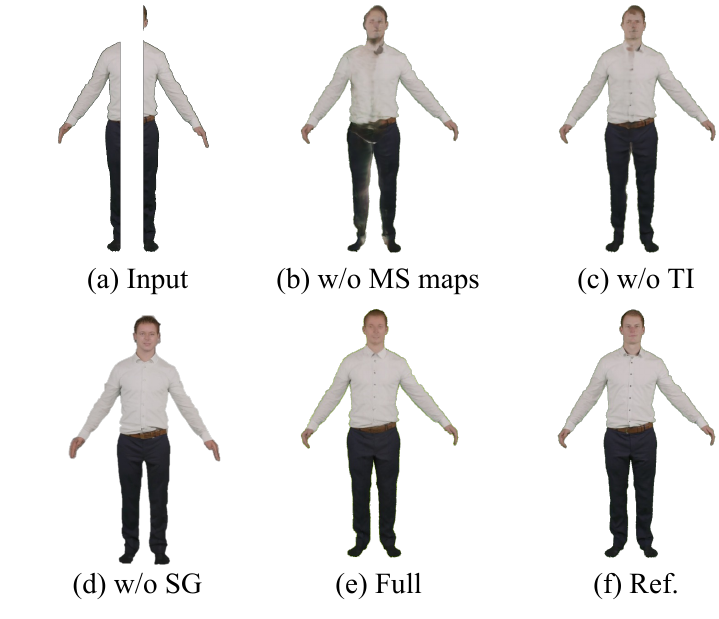}
    \caption{We visualize the effect of each component on the PeopleSnapshot sequence with synthetic occlusions.}
    \label{fig:ablation}
\end{figure}

\paragraph{Qualitative Analysis.}
Figure~\ref{fig:ablation} visualizes the effect of each component on reconstruction quality. Without multi-scale feature maps, the model fails to effectively interpolate information across occluded areas, resulting in overly smooth textures that lack fine-grained details. Without textual inversion, the diffusion model can still complete occluded regions but tends to generate content that deviates from the subject's actual appearance on some frames, for instance, producing clothing with incorrect colors or patterns, leading to noticeable visual inconsistencies. This observation directly validates the importance of subject-level tokenization in preserving identity during the inpainting process. Without semantic guidance, the model struggles to maintain part-level consistency, particularly for semantically meaningful regions such as the face, where anatomical coherence is crucial.

%% file: sec/5_Conclusion.tex
\section{Conclusion}
\label{sec:conclusion}

We have presented InpaintHuman, a method for reconstructing complete and animatable 3D human avatars from occluded monocular videos. Our approach addresses the challenge of missing observations through two synergistic components: a multi-scale UV-parameterized canonical representation enabling robust feature interpolation across partially occluded regions, and an identity-preserving diffusion inpainting module leveraging personalized generative priors for subject-specific completion. By employing direct pixel-level supervision rather than stochastic SDS-based optimization, our method achieves improved reconstruction quality while maintaining identity consistency.

Experiments on both synthetic and real-world benchmarks demonstrate competitive performance compared to state-of-the-art methods. Ablation studies validate the effectiveness of multi-scale feature design for occlusion robustness and subject-level tokenization for identity preservation. We hope this work provides useful insights for human digitization under challenging real-world conditions.

\paragraph{Limitations and Future Work.}
Several limitations warrant future investigation. First, our method relies on SMPL parameters from off-the-shelf estimators; severe occlusions may cause inaccurate poses that propagate errors. Second, for completely unobserved regions, our diffusion module may generate plausible but not necessarily ground-truth-accurate content, an inherent limitation of generative approaches.